\documentclass{article}

\usepackage{microtype}
\usepackage{graphicx}
\usepackage{float}
\usepackage{subcaption}
\usepackage{booktabs} % for professional tables

\usepackage{hyperref}
\usepackage{amsmath}
\usepackage{amsfonts}
\usepackage{adjustbox}
\usepackage[accepted]{icml2021}

\icmltitlerunning{Learning to Infer Belief Embedded Communication}

\begin{document}

\twocolumn[
\icmltitle{Learning to Infer Belief Embedded Communication
}
% \icmltitle{Multi-Agent Reinforcement Learning via Learning Motivation}
% \icmltitle{Learn to Communicate with Message Confidence Uncertainty}
% It is OKAY to include author information, even for blind
% submissions: the style file will automatically remove it for you
% unless you've provided the [accepted] option to the icml2021
% package.

% List of affiliations: The first argument should be a (short)
% identifier you will use later to specify author affiliations
% Academic affiliations should list Department, University, City, Region, Country
% Industry affiliations should list Company, City, Region, Country

% You can specify symbols, otherwise they are numbered in order.
% Ideally, you should not use this facility. Affiliations will be numbered
% in order of appearance and this is the preferred way.
%\icmlsetsymbol{equal}{*}

\begin{icmlauthorlist}
\icmlauthor{Guo Ye}{nw,zeb}
\icmlauthor{Han Liu}{nw}
\icmlauthor{Biswa Sengupta}{zeb}
\end{icmlauthorlist}

\icmlaffiliation{nw}{Department of Computer Science, Department of Statistics, Northwestern University, USA}
\icmlaffiliation{zeb}{Zebra Technologies, London, UK}
\icmlcorrespondingauthor{Biswa Sengupta}{biswa.sengupta@zebra.com}

% You may provide any keywords that you
% find helpful for describing your paper; these are used to populate
% the "keywords" metadata in the PDF but will not be shown in the document
\icmlkeywords{Machine Learning, ICML}

\vskip 0.3in
]

% this must go after the closing bracket ] following \twocolumn[ ...

% This command actually creates the footnote in the first column
% listing the affiliations and the copyright notice.
% The command takes one argument, which is text to display at the start of the footnote.
% The \icmlEqualContribution command is standard text for equal contribution.
% Remove it (just {}) if you do not need this facility.

\printAffiliationsAndNotice{}  % leave blank if no need to mention equal contribution
%\printAffiliationsAndNotice{\icmlEqualContribution} % otherwise use the standard text.

\begin{abstract}
In multi-agent collaboration problems with communication, an agent's ability to encode their intention and interpret other agents' strategies is critical for planning their future actions. This paper introduces a novel algorithm called Intention Embedded Communication (IEC) to mimic an agent's language learning ability. IEC contains a perception module for decoding other agents' intentions in response to their past actions. It also includes a language generation module for learning implicit grammar during communication with two or more agents. Such grammar, by construction, should be compact for efficient communication. Both modules undergo conjoint evolution - similar to an infant's babbling that enables it to learn a language of choice by trial and error. We utilised three multi-agent environments, namely predator/prey, traffic junction and level-based foraging and illustrate that such a co-evolution enables us to learn much quicker (50 \%) than state-of-the-art algorithms like MADDPG. Ablation studies further show that disabling the inferring belief module, communication module, and the hidden states reduces the model performance by 38\%, 60\% and 30\%, respectively. Hence, we suggest that modelling other agents' behaviour accelerates another agent to learn grammar and develop a language to communicate efficiently. 
We evaluate our method on a set of cooperative scenarios and show its superior performance to other multi-agent baselines. We also demonstrate that it is essential for agents to reason about others' states and learn this ability by continuous communication.

%Modeling the uncertainty into the communicating process between agents is more adaptive to real world scenarios. In this paper, we propose an algorithm that formulate the confidence of the information with a distribution that can affiliate the efficiency of the message exploitation.

% We show that our method can outperform other multi-agent baselines with relatively fewer communication cost. The method can be used in multi-agent warehouse scenarios where robust network may not be guaranteed.
\end{abstract}

\begin{figure}[th]
    \centering
    \includegraphics[scale=0.5]{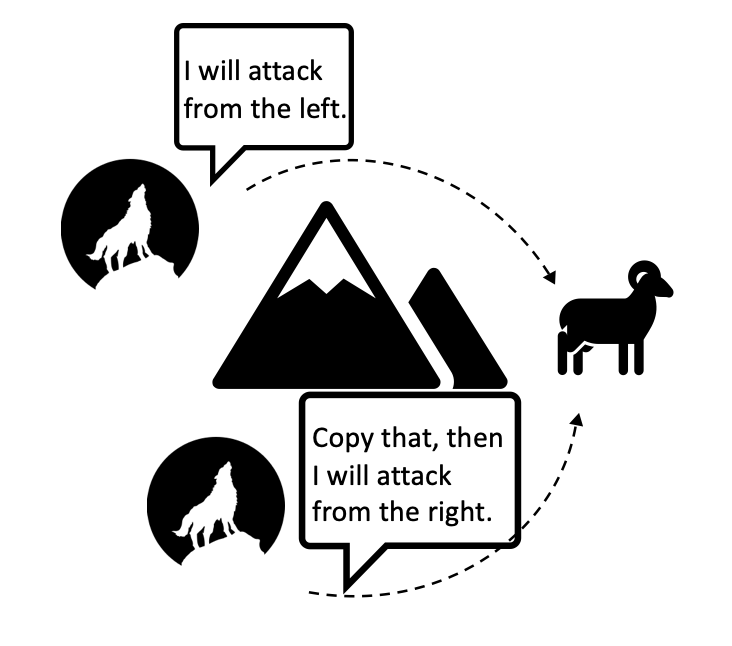}
    \caption{Motivation: To successfully achieve a cooperation based hunting task, the wolves are required to 1) compose their howl stating their strategies to surround the lamb and 2) for other wolves to correctly understand such a strategy being communicated by members of its pack. }
    \label{fig:recurisive_reasoning_illustration}
    \vspace{-0.5cm}
\end{figure}
\vspace{-0.5cm}

\section{Introduction}
Multi-agent reinforcement learning (MARL) has drawn significant attention due to its wide applications in games \cite{mnih2015human, silver2016mastering}. But when moving into a multi-agent environment, agents get rewards not only depending on the state and its action but also other agents' actions \cite{tian2019regularized}. Agents' changing policies bring out the non-stationarities of the multi-agent environment; recent MARL methods mainly utilize opponent modeling \cite{zintgraf2021deep}, communication \cite{sukhbaatar2016learning,singh2018learning} and experience sharing \cite{christianos2021scaling} to address such credit assignment problems \cite{christianos2021scaling,papoudakis2020variational}.
The multi-agent environments can be grouped into three groups, cooperative, competitive and hybrid. In this work, we mainly focus on cooperative tasks. Under such a setting, efficient and correct interpretation of others' intentions is crucial in provisioning resources and achieving the ultimate goal. Communicating with other agents is one way of inferring such intentions.

In nature, social animals can quickly establish an adequate understanding of each other's mental state and choose the actions based on that \cite{tian2020learning} in collaborative tasks. That's one of their essential skills to live. 
Several applications \cite{lazaridou2016multi, mordatch2018emergence} introduce the emergence of natural language under such a multi-agent cooperation environment. The ability to infer other agent's implicit motivation and maintain a mental state about their strategies refers to machine theory of mind \cite{rabinowitz2018machine}. 
This work proposes a framework closer to a real situation in which agents learn to reason others' beliefs by continuous exchanges, recursively (Figure \ref{fig:recurisive_reasoning_illustration}). Imagine a group of wolves hunting lambs, each of the team member need to understand others' howl and, based on its situation, draw up a response. To mimic this, we utilise variational autoencoders as a comprehension module responsible for translating the received message.  The agent will first use it to reason the sender's strategy and characteristics \cite{tian2019regularized}; in our work, the agent's following observation and reward is a response to the sender's strategy in the prior state. On one side, this information will be added to its own observation for decision making. On another side, there is a module parametrised by MLP responsible for generating the response on account of its state and inference on others. 
We found that with the mechanism we defined above, the agents can reach higher performance than other state-of-art methods. 
\begin{figure*}[ht]
    \centering
    \includegraphics[width=\textwidth, scale = 0.8]{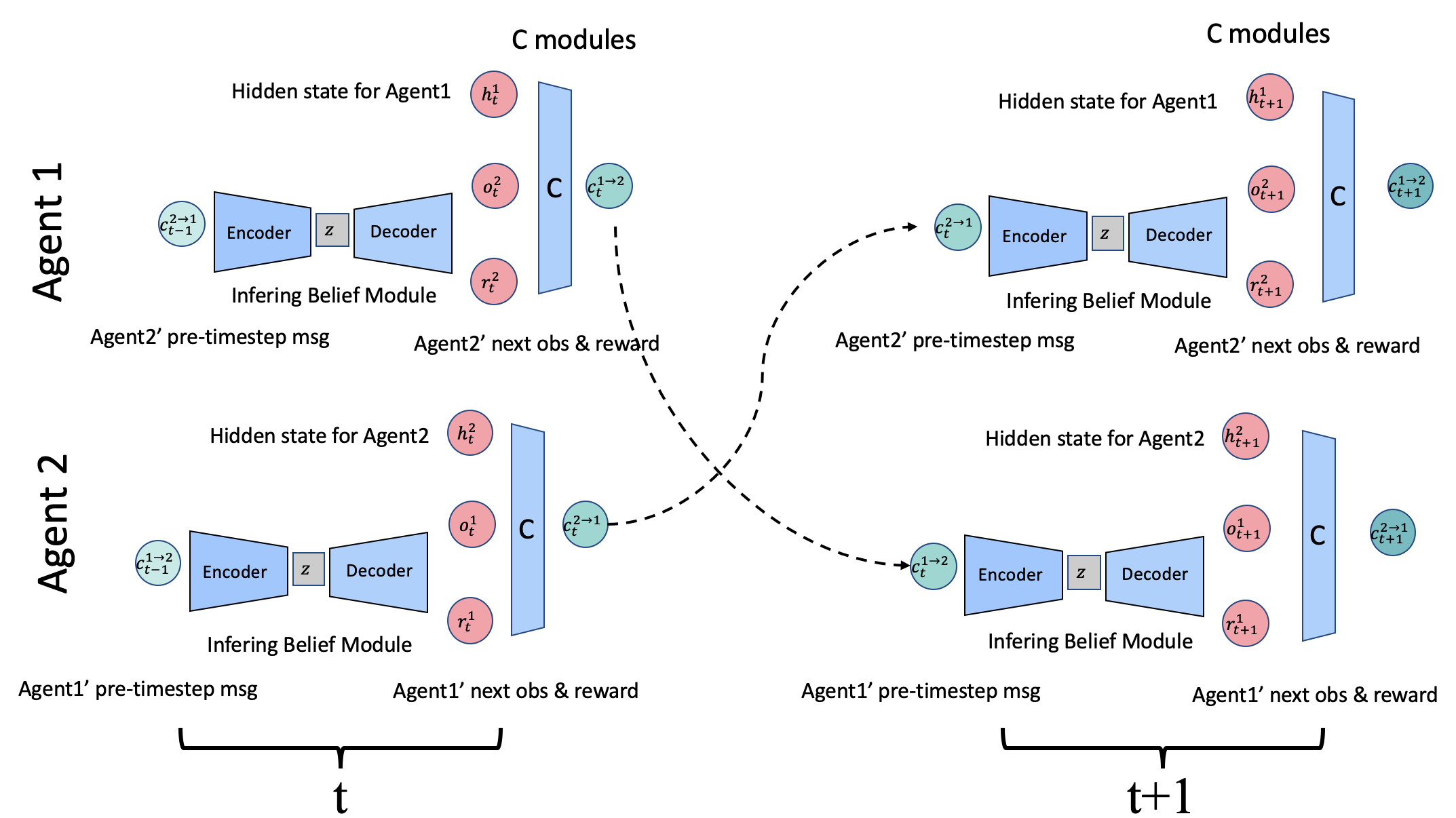}
    \caption{The communication architecture under a multi-agent setting. Suppose we have $N$ agents, each of them at time $t$ receives the observation $o_t$ and the communication message $c_t$. The message is reconstructed by a VAE component which takes in and sums up all the other agents' hidden state. }
    \vspace{-0.1cm}
    \label{fig:architecture}
\end{figure*}

Specifically, our architecture contains three functional modules (see Figure \ref{fig:architecture} and \ref{fig:variable_flow}). First, each agent has an \textbf{inferring beliefs module (IBM)}, maintaining its interpretation and modelling others under uncertainty. And a \textbf{communication module} (C module) learns how to compress and encode an agent's observation and belief into compact codes. Finally, a \textbf{policy module} makes decisions based on the interpretation of the ongoing communication.
We put forward three  contributions: 1) We propose a generative module to infer other agents' beliefs; 2) We give the agents the ability to learn how to generate effective (concise) messages; 3) We showed that our method could learn to reach higher performance with all these functionalities.

\section{Related Work}
\subsection{Modeling Other Agents}
Opponent modelling is a popular and promising research direction in multi-agent reinforcement learning (MARL). The unstable nature of the shared MARL environment has encouraged works to integrate other agents' models into the controlled agent's policy. By including a belief in the opponent's behaviour, the uncertainty of the environment is partially relieved during training. 

He et al. \cite{he2016opponent} propose DRON, building a reinforcement learning framework based on DQN, which jointly learns the policy and the opponent modelling conditioned on observations. 
Zhang et al. \cite{zhang2020robust} takes the model uncertainty into account while trying to make the MARL algorithm more robust. Chrisianos et al. \cite{christianos2021scaling} propose SePS where they model other agents' resulting observation and reward. They do so by pre-training a clustering mechanism. MoT (Machine theory of Mind) \cite{rabinowitz2018machine} creates two latent variables called agent character $m$ and mental state $m_t$ to build a belief on other agents. They use past trajectory for updating the character and current trajectory for mental state. Zintgraf et al. \cite{zintgraf2021deep} borrow this machine theory of mind idea and pass it to other agents who use Bayesian inference to obtain the predicted next action. Previous methods require access to the opponent's information, Papoudakis et al. \cite{papoudakis2020variational} propose that the agent can reason about others by just accessing its observation.

\subsection{Multi-agent RL based on Communication}
For multi-agent reinforcement learning, each agent only has partial observation of the environment, leading to difficulty for learning. It's naturally possible to introduce a communication channel to provide additional information for the agents when they're learning cooperative or competitive tasks. This leads us to two questions. 1) How can we construct the message? and 2) Who to communicate?   For the first question, CommNet \cite{sukhbaatar2016learning} propose a model that can keep a hidden state for each cooperative agent, and the message is the average value of all other agents' hidden state. The fully-connected network brings lots of computational burden, especially as the number of agents increases, the length of the message could increase exponentially. Also, it becomes more difficult for an agent to filter the informative values among accumulated messages. IC3Net \cite{singh2018learning} add a gate mechanism on the top of CommNet, where another binary action is then generated for determining whether the agent would like to send the message. Vertex Attention Interaction Network (VAIN) \cite{hoshen2017vain} adds an attentional architecture for multi-agent predictive modeling. Instead of averaging the messages, they put weights in front of each vector.

\section{Background}
\subsection{Markov Games}
We formulate our environment as a Markov game \cite{shapley1953stochastic,littman1994markov,ye2020collision} with $N$ agents defined by the following tuple: $$\mathcal{M} = (S,O_{1}\ldots O_{N},A_{1}\ldots A_{N},T,R_{1}\ldots R_{N}).$$ 
Here $N$ is the total number of agents. 
$S$ denotes the states of all agents. $O_{1}\ldots O_{N},A_{1}\ldots A_{N}$ are the sets observations and actions for each agent. $A_i$ is action of Agent $i$  sampled from a stochastic policy $\pi_i$ and the next state is generated by the state transition function $T: S \times A_{1}\times \ldots \times A_{N} \to S$. At each step, every agent gets a reward according to the state and corresponding action $r_i:S \times A_i \times \ldots \times A_N \to \mathbb{R}$ along with an observation of the system state $o_i: S \to O$. The policy of agent $i$ is $\pi_i$.  
The objective for the $i$-th agent is to learn a policy that maximizes the cumulative discounted rewards 
\begin{eqnarray}
\mathcal{R}(i):= \mathbb{E} \Big[ \sum_{t=0}^{T} \gamma^{t} r_i^{(t)})\Big], \label{eq::MGobj}
\end{eqnarray}
where $\gamma \in (0,1)$ is the discount factor and $r_i^{(t)}$ is the reward received at the $t$-th step.

\subsection{Variational Autoencoders}
 Variational auto-encoder (VAE) is a generative model. It can learn a density function $p(z|x)$, where $z$ is a normally distributed latent variable, $x$ is the given input from dataset $X$. 
 The true posterior distribution $p_\theta(\mathbf{z}|\mathbf{x})$ is unknown, 
%  it contains an encoder which learns a variational parametric distribution denoted $q_{\theta}(z|x)$ parametrized by $\theta$. The KL-divergence between the learned distribution and the true posterior is 
% \begin{equation*}
%     D_{KL}(q_{\theta}(z|x)||p(z|x)) = log p(x) - \mathbb{E}_{z}
% \end{equation*}
% the recognition model is parametrized by $\phi$, which encodes the data to the approximate posterior.
 so the goal of VAE is to approximate a distribution $q_\phi(\mathbf{z}|\mathbf{x})$ which is parametrized by multilayer perceptrons (MLPs). VAEs also use a MLP to approximate $p_\theta(x|z)$. The KL-divergence between two is following: 
 \begin{equation*}
 \begin{aligned}
     D_{KL}(q_\phi(\mathbf{z}|\mathbf{x} \| p_\theta(\mathbf{z}|\mathbf{x}) = log p_\theta(\mathbf{x}) 
     + D_{KL}(q_\phi(\mathbf{z}|\mathbf{x}) \parallel p_\theta(\mathbf{z})) \\
     - \mathbf{E}_{z \sim q_\phi(\mathbf{z}|\mathbf{x}) } log p_\theta (\mathbf{x} | \mathbf{z}),
 \end{aligned}
 \end{equation*}

which is equal to 
 \begin{align*}
    log p_\theta(\mathbf{x}) - D_{KL}(q_\phi(\mathbf{z}|\mathbf{x} \| p_\theta(\mathbf{z}|\mathbf{x}) = \\
      \mathbf{E}_{z \sim q_\phi(\mathbf{z}|\mathbf{x}) } log p_\theta (\mathbf{x} | \mathbf{z}) - D_{KL}(q_\phi(\mathbf{z}|\mathbf{x})  \parallel p_\theta(\mathbf{z})).  
 \end{align*}

The right side of the equation is called the evidence lower bound (ELBO). Since the second term of the left side is equal or larger than zero, we can get the following inequation:

\begin{equation*}
        log p_\theta(\mathbf{x}) \ge \\
      \mathbf{E}_{z \sim q_\phi(\mathbf{z}|\mathbf{x}) } log p_\theta (\mathbf{x} | \mathbf{z}) - D_{KL}(q_\phi(\mathbf{z}|\mathbf{x})  \parallel p_\theta(\mathbf{z})),  
\end{equation*}
so minimising the loss function is equivalent to maximize the ELBO. The density forms of $p(\mathbf{z})$ and $q_\phi(\mathbf{z}|\mathbf{x})$ are chosen to be multivariate Gaussian distribution. The empirical objective of the VAE is 
\begin{align*}
\begin{gathered}
  L_{VAE}(\theta,\phi) = - \mathbf{E}_{z \sim q_\phi(\mathbf{z}|\mathbf{x}) } log p_\theta (\mathbf{x} | \mathbf{z}) + D_{KL}(q_\phi(\mathbf{z}|\mathbf{x})  \parallel p_\theta(\mathbf{z})), \\
  \theta^*,\phi^* = arg \min_{\theta,\phi} L_{VAE}.
\end{gathered}
\end{align*}

The recognition and generative models are parametrised using multilayer perceptrons (MLPs).

% \section{Opponent Modeling Embedded Message}
% \section{Reasoning Embedded Message Recursively}
% \section{Intention Embedded Communication}\label{IEC}
\section{Belief Embedded Communication}\label{IEC}

%recursive reasoning

Most opponent modeling methods require direct access of other agents' information \cite{papoudakis2020variational} or history of trajectories \cite{he2016opponent,raileanu2018modeling,li2018dynamic}. To lean close to real-life situations, in our hypothesis, agents can only build their beliefs by conversing with each other.
Hence, we want to enable every single agent to comprehend and send the latent messages. Via the information exchange, they also learn to recursively reason, wherein they are making decisions based on the knowledge of what the opponent would do based on how they would behave \cite{wen2019probabilistic}.

% \subsection{Intention Belief Module}
\subsection{Inferring Belief Module}
Just as the human brain has an area for language processing to understand other agent's conversation, we construct a similar functional module in our framework. 
% We propose a multi-agent communication method with message integrated uncertainty. 
Essentially, we utilise a VAE (Variational Auto Encoder) to represent the belief module and model the agents' motivation. The message embeds other agents' future states including represented by following observation and rewards. Using this module, an agent can decode intentions from other agents' embedded messages in a decentralised manner instead of direct access to other agents' states.

In addition, every human understands language in the real world using their own implicit model of comprehension. So unlike prior methods, the trained VAE model is not shared by all the agents. We maintain an IBM (Inferring Belief Module) for each agent, i.e., they interpret the received message.

The encoder $q$ and decoder $p$ of VAE are parametrized by $\phi$ and $\theta$ respectively. The input of the encoder are others' embedded message $c_t$ at timestep $t$ which is a $C \times (N-1)$ shape tensor. $C$ is the length of communication vector, $N$ is the total number of agents. We assume each agent can faithfully receive others message without any information loss. The output of the encoder is a $m$-dimensional Gaussian distribution $N(z;\mu,\sigma^2)$. The decoder predicts the $p(x_{t+1}|z)$, where $x_{t+1}=(o_{t+1},r_{t+1})$ is the next observation and reward.

The agents' beliefs can be projected to the latent space $Z$ by $q_\phi(z|c_t)$ and $p_\theta(z|x_{t+1})$. Hence the objective is to optimize $D_{KL}(q_\phi(z|c_t)\parallel p_\theta(z|x_{t+1}))$. Since $p(x_{t+1})$ is intractable, it's equal to optimize evidence lower bound (ELBO) which we derive as following:
\begin{equation}
\begin{split}
    log p(x_{t+1}) \geq \mathbb{E}_{q_{\theta} \sim (z|c_{t})}[log p_\theta(x_{t+1}|z) \\
    -D_{KL}(q_{\theta}(z|c) \parallel p(z))]
\end{split}
\end{equation}

We collect the trajectories and messages for certain period of time. The IBM of each agent is trained independently. While training, the trajectories of all agents are collected and saved to a buffer, for every 50 epochs, we train VAEs 10 times.
% The reconstruction term of the ELBO factories as:

% \begin{align}
%     log p_u(tr|z) = log p_u(r_t,o_{t+1}|a_t,o_t,z)p(a_t,o_t|z)= \\log p_u(r_t,o_{t+1}|a_t,o_t,z) + log p(a_t,o_t|z)
% \end{align}

\subsection{Policy and Message Generation Module}\label{subsec:message_generation_module}
\begin{figure}
    \centering
    \includegraphics[scale=0.35]{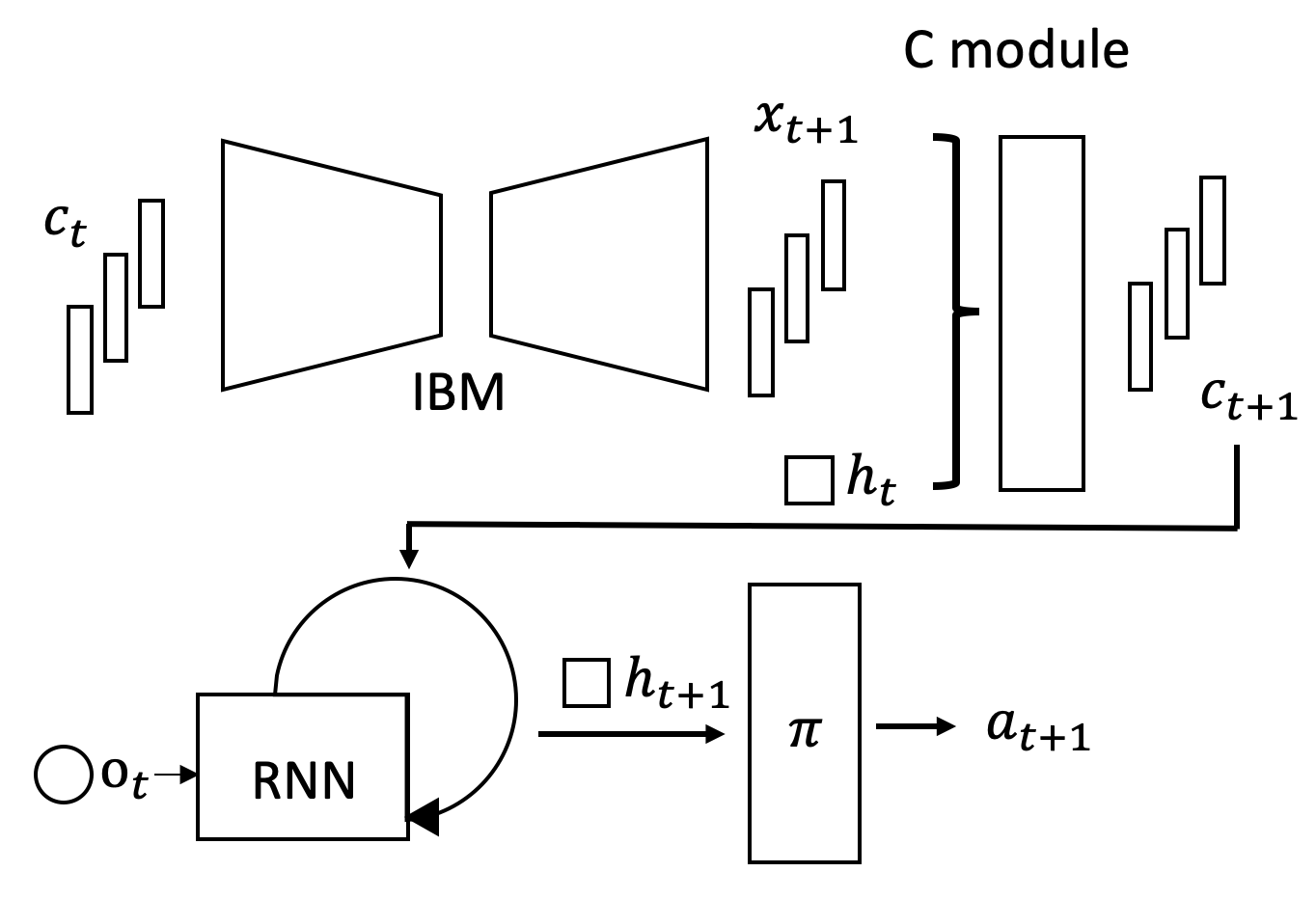}
    \caption{The variable flow graph of Message and policy generation module. The strip shaped bars and squares are the vectors flow in and out the modules. The message pass through the IBM (Intention Belief Module) to get the predictions of other agent's intention named $x_{t+1}$. C module then utilize the predictions and agent's own hidden state to generate next timestep's communication vector $c_{t+1}$. The policy module add in observation and passing through a RNN and policy network outputing action $a_{t+1}$.}
    \label{fig:variable_flow}
\end{figure}

For agents, the policy module takes in the observation from environments and combines the beliefs of others' intentions predicted by previous IBM. It outputs the message delivered to others and subsequent actions agents will take.

The new generated communication vector is continuous, it's aggregated by all the IBM predictions that represent the agent's belief and agent's own hidden state (see Figure \ref{fig:variable_flow}). For agent $j$, $c^j_{t+1}$ is the generated message to others at timestep $t$ shown as following:

\begin{equation*}
\begin{aligned}
    c^j_{t+1} = \frac{1}{N-1} C (\sum_{j=1}^{N-1} x_{t+1}^j + h_{t}) \\
    % x_{t+1}^j = IBM(c^j_{t})
\end{aligned}
\end{equation*}
\begin{equation*}
  x_{t+1}^j = IBM(c^j_{t})
\end{equation*}
\begin{equation*}
    h_{t+1} = RNN(o_t+c_{t+1}).
\end{equation*}

The message generation module $C$ is parametrised by a fully connected neural network. Previous IBM(Inferring Belief Module) decodes $x_{t+1}^j$. There is no importance factor multiplied by each vector. We also prepare an RNN module to keep track of the agent's hidden state like \cite{singh2018learning,sukhbaatar2016learning}. Next timestep's hidden state $h_{t+1}$ is generated by this recurrent module where the input is current step's observation $o_t$ and the new generated communication $c_{t+1}$
The next action is generated using policy network 
implemented as fully-connected neural network $a^j_{t} = \pi(h_t)$.

\subsection{Implementation}

\begin{algorithm}[!t]
\begin{algorithmic}[1]
  \REQUIRE Total episode number $E$, Duration of each episode $T$, Number of agents $N$. VAE training interval $I$.
  \STATE Initialize policy $\pi_0$, VAE models' parameter $\theta, \phi$, RNN module. \\
  Initialize memory buffer $\mathbf{B}$ \\
  Initialize IBM's $N$ replay buffer $\mathbf{D}=\{D^0,\cdots, D^N\}$
    \FOR{$k =0, \ldots, E-1$}
        \STATE Play an episode using current policy $\pi_k$.
        \STATE add trajectory $\tau_k$ into memory buffer $\mathbf{B}$
        \FOR{$j=0, \ldots, N$}
            \STATE add $c^{-j}$, $s^{-j}$ and $r^{-j}$ into replay buffer $D^j$
        \ENDFOR
        \STATE Update policy \(\pi_{k+1} = \pi_k + \beta \nabla_{\theta}{\hat{J}_\pi(\pi_k)}\) and RNN using the data in memory buffer $\mathbf{B}$. 
        \IF {k is divisible by $I$}
            \FOR{$j=0, \ldots, N$}
                \STATE Update $\theta,\phi$ using samples from $D^j$
            \ENDFOR
        \ENDIF
    \ENDFOR
\STATE return optimal $\pi^*,\theta^*,\phi^*$
\end{algorithmic}
\caption{Intention Embedded Communication}
\label{algorithm:IEC}
\end{algorithm}

% \begin{algorithm}[tb]
%   \caption{Bubble Sort}
%   \label{alg:example}
% \begin{algorithmic}
%   \STATE {\bfseries Input:} data $x_i$, size $m$
%   \REPEAT
%   \STATE Initialize $noChange = true$.
%   \FOR{$i=1$ {\bfseries to} $m-1$}
%   \IF{$x_i > x_{i+1}$}
%   \STATE Swap $x_i$ and $x_{i+1}$
%   \STATE $noChange = false$
%   \ENDIF
%   \ENDFOR
%   \UNTIL{$noChange$ is $true$}
% \end{algorithmic}
% \end{algorithm}

\begin{figure*}[t]
    \centering
    \begin{subfigure}{.32\textwidth}
        \includegraphics[scale=0.8]{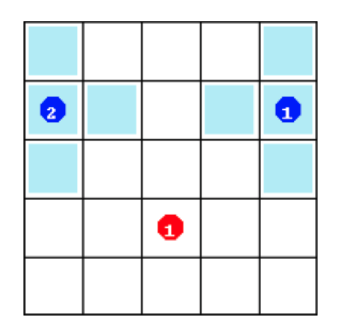}
        \caption{ PredatorPrey}
    \end{subfigure}
    \begin{subfigure}{.32\textwidth}
        \includegraphics[scale=0.45]{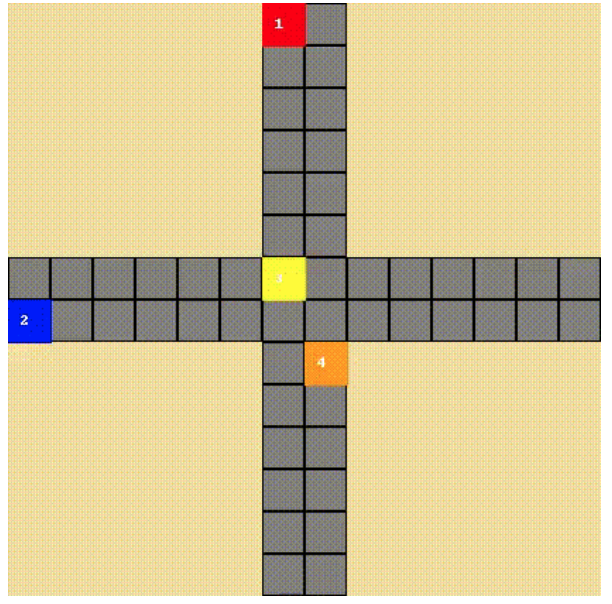}
        \caption{ TrafficJunction}
    \end{subfigure}
    \begin{subfigure}{.32\textwidth}
        \includegraphics[scale=0.73]{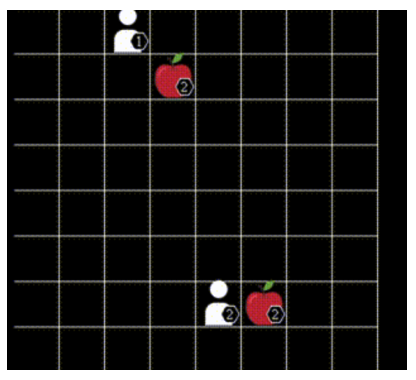}
        \caption{ Level-based Foraging}
    \end{subfigure}
    \caption{Environments used in our experiment. Left is PredatorPrey Env. The blue circles with numbers represent the predators and the red one is the prey. The azure blocks around the predator illustrate their (limited) visual inputs. The default grid size is $5 \times 5$. In our setting, the grid size, number of agents and vision are changeable according to the different difficulty settings. The center is TrafficJunction Env. Each agent represented by colored circles need to pass the crossroads without collision. Their actions are limited to \textit{gas} and \textit{brake}. The right is Level-based Foraging, the largest difference of this environment and PredatorPrey is that it requires a level verification while collecting the targets.}
    \label{fig:envs}
\end{figure*}

% generate a normal distribution with parameters $\mu, \sigma$, then the message is passed to other agents, and they use a corresponding decoder to retrieve the information for the next timestep's decision making.

As Algorithm \ref{algorithm:IEC} shows, we first initialise a policy $\pi_0$ and $N$ belief modules, each representing specific agent's comprehension of the received message.
We train the policy learning module and inferring belief module (IBM), separately. Since IBM is trained along with the policy, as the agents' policy evolves, the comprehension ability updates at set intervals. Specifically, for every $I$ episode, our method collects this period of experience into buffer $\mathbf{D}=\{D^0,\cdots, D^N\}$. For each agent's buffer $D^j$, it contains others' message $c^{-j}$, next state $s^{-j}$ and reward $r^{-j}$ for training. In the next update iteration, the buffers are emptied because the context between agents is changed. Memory buffer $\mathbf{B}$ collects all the interactions for each step, including the rewards, then the pairs are used to train the policy $\pi$ like standard reinforcement learning methods. By the end, we anticipate that all agents can learn to hear (inferring other's messages correctly) and speak (generate informative messages) in an efficient way.
% The encoder $\xi_\theta$ takes in the messages $c^i_t$ from other agents and outputs embedding $z^i_t$, the decoder recover the others' observations $o^{-i}_{t}$ and rewards $r^{-i}_{t}$. See algorithm for details.
We follow the centralized training with decentralized execution (CDTE) paradigm.
%TODO
%A graph about the general architecture containing VAE.(refering to RIAL/DIAL)
% \begin{align*}
%     c^{t} = Decoder(\mu, \sigma) \\
%     s^{t} = o^{t} + c^{t}
% \end{align*}

\section{Experiment}
The following sections present the experiments where we evaluate our approach in multiple multi-agent environments including PredatorPrey \cite{singh2018learning}, TrafficJunction \cite{magym} and Level-base Foraging \cite{papoudakis2021benchmarking} showing in Figure \ref{fig:envs}. These environments require effective coordination between agents. We exhaustively compare our algorithm (IEC) with other state-of-art multi-agent methods in each environment and probe into the different configurations and parameters that may affect the performance. 
All results presented are averaged over five independent seeds. Our algorithm's network architecture is kept the same for all runs. The learning rate is $1 \times 10^{-3}$ and batch size is $500$. The buffer size for VAE is $4 \times 10^4$; we trained the VAE every 50 episodes.

\subsection{Baselines}\label{subsection:baselines}
We compare our method with following popular multi-agent methods as baselines.

\textbf{VDN:} Value-decomposition Networks \cite{sunehag2017value} factorize the joint value function into $N$ Q-functions for $N$ agents. Each agent's value function only relies on its local trajectory. \\
\textbf{QMIX:} QMIX \cite{rashid2018qmix} is a value-based off-policy algorithm that extends the VDN with a constraint that enforces the monotonic improvement. \\
\textbf{MADDPG:} Multi-Agent Deep Deterministic Policy Gradient \cite{lowe2017multi} use trajectories of all agents to learn a centralized critic. At execution phase, the actors act in decentralized manner. \\
% \textbf{iDQN:} iDQN use a \\

% \subsection{Environment Setting}
% We test our method on three simulated environments showing in Figure\ref{fig:envs}.

\subsection{PredatorPrey}
 In the PredatorPrey environment, $N$ predators are assigned to capture the prey. The setting is that only after all predators reach the prey's location can the reward be given to them. The observation of a given predator is its concatenation of the nearby grid. The actions are discrete options (\texttt{Forward, Backward, Left, Right, Stale}) for all agents. 
%  In the first experiment, we locked the prey's position; in latter ones, the prey can escape. 
 The maximum number of steps for each episode is 20. 
 %The reward is assigned to predators when both of them reach the prey's location. 

\begin{figure*}
    \centering
    \vspace{-0.2cm}
    \begin{subfigure}{.32\textwidth}
    \includegraphics[scale=0.43]{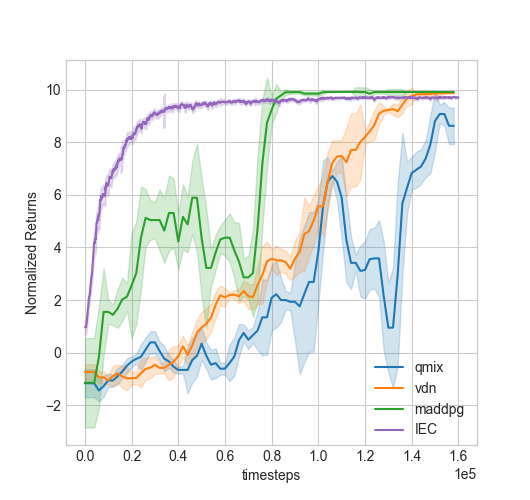}
    \caption{PredatorPrey}
    \label{PredatorPrey}
    \end{subfigure}
    \begin{subfigure}{.32\textwidth}
    \includegraphics[scale=0.43]{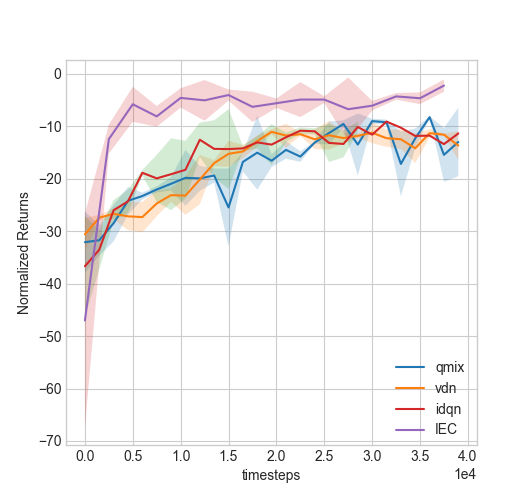}
    \caption{TrafficJunction}
    \label{TrafficJunction}
    \end{subfigure}
    \begin{subfigure}{.32\textwidth}
    \includegraphics[scale=0.43]{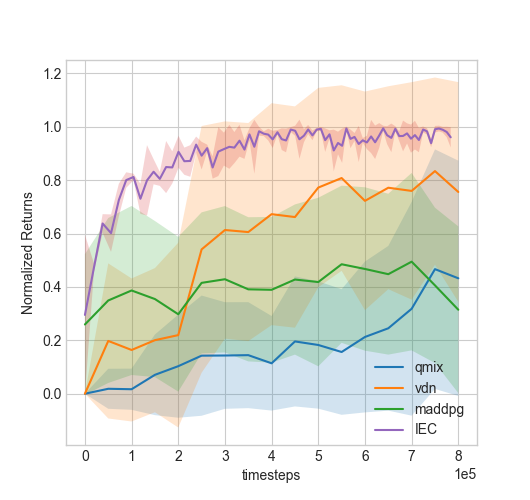}
    \caption{Level-based Foraging}
    \label{Level-based Foraging}
    \end{subfigure}
    \caption{Performance of IEC and other baselines listed in \ref{subsection:baselines} during the training protocol. The y-axis is normalized returns and x-axis is the total time steps the algorithm takes. As shown in three subplots, IEC can learn the task faster and achieve higher or equal returns by the end.}
    \label{fig:baselines}
\end{figure*}

%Compare Baselines
The first experiment we conducted is a simple cooperative setting with $5 \times 5$ grid as a base map. For every episode, 2 predators and 1 prey are randomly allocated in the grid world. When all the predators reach the prey's location, they are awarded 5 points. We sum up all predators's rewards as an indicator of performance. We run experiments for 5 trials with independent seeds. The learning rate for our algorithm and other baselines is 0.001. As shown in Figure \ref{PredatorPrey}, our algorithm converges speedily compared to others. It uses around half of the interactions compared to the second-best MADDPG. 

%Diff env difficulty
We then tested our algorithm on various environment settings. Specifically, we follow most settings from \cite{singh2018learning} to construct three levels of difficulty with different agents numbers, grid size and agents' vision. The details can be found in Table \ref{tab:Diff_PP_Env_Settings}. The communication vector length in the three environments is the default 128 lengths. 
\begin{table}[]
    \centering
    \begin{adjustbox}{width=\columnwidth,center}
    \begin{tabular}{c|c|c|c}
                & Agents Num & Grid Size & Agent's Vision \\
    Easy Version & 2 & 5 & 0 \\
    Medium Version & 4 & 10 & 1 \\
    Hard Version & 10 & 20 & 1 \\
    \end{tabular}
    \end{adjustbox}
    \caption{Settings of different versions of the PredatorPrey environment}
    \label{tab:Diff_PP_Env_Settings}
\end{table}
\begin{figure}[h]
    \centering
    \includegraphics[scale=0.55]{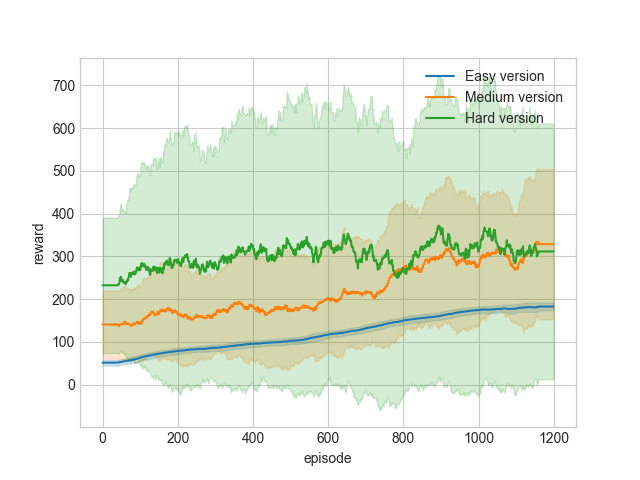}
    \vspace{-0.7cm}
    \caption{The learning curves of IEC on different difficulty levels of the PredatorPrey Env. It shows that as environment becomes more complicated, the variance notably increases.}
    \label{fig:PP_diff_env}
\end{figure}
As Figure \ref{PredatorPrey} shows, when the environment becomes more difficult for the agents, the variance enlarges, since in our setting, we assign the rewards to agents only when all the agents reach the prey's location. As the number of agents increase and the grid size expands, it's challenging for them to finish the task in limited time steps.

%Diff comm size
We conduct several experiments for investigating the influence of the message size. Our communication consists of the hidden state, and the agent's beliefs of others. The message generation module concatenates these two together and forwards them to a trainable neural network for the final outputs. The length of the message is adjustable, so in the following experiment, we probe into the influence of the length of the different bits.
We conduct experiments with 32, 64 and 128 bits of message and show their training process.
As observed in the left plot of Figure \ref{fig:diff_bits}, the length of the communication vector has a significant influence on policy optimisation. Longer bits can carry more encoded information and facilitates communication between agents.

\subsection{Traffic Junction}
The traffic junction environment we use references to \cite{singh2018learning} and \cite{magym}. In this environment, cars move across intersections in which they need to avoid collisions with each other. There are adjustable routes users can set. The number of routes and agents define the three levels of difficulty named easy, medium and hard version. 

% The mixed environment setting is that the reward will assign to agents only when all of them have reached the environment. It's not explicitly signal to encourage agents to cooperate.

%Baselines
We first compared our method with other baselines. As Figure \ref{TrafficJunction} shows, agents trained by our method can learn to cooperate at a much faster speed and also furnish higher final returns. The other methods are nearly of the same performance. The TrafficJunction environment requires agents to utilise all local observations to get the other agent's position and intention. This setting brings a great advantage for our method because the belief and cooperation information can be encoded into a message. Under our framework, the negotiation can be reached by mutual recursive reasoning to avoid a collision.

\begin{figure}[h]
    \centering
    \includegraphics[scale=0.55]{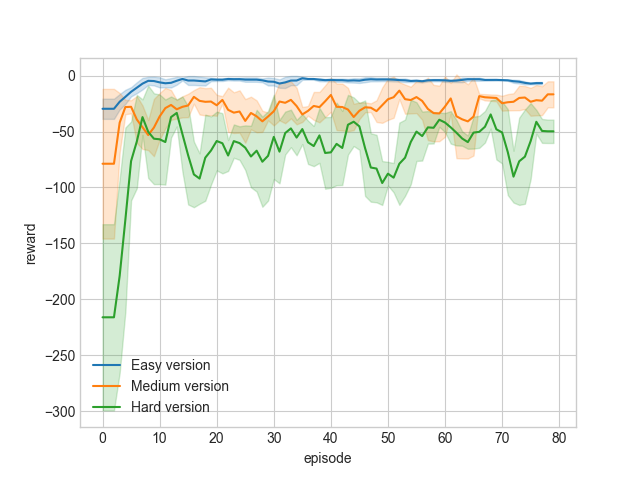}
    \vspace{-0.7cm}
    \caption{The learning curves of IEC on different difficulty level of TrafficJunction Env.}
    \label{fig:traffic_junction_diff_env}
\end{figure}

\begin{figure*}[h]
    \centering
    \includegraphics[scale=0.45]{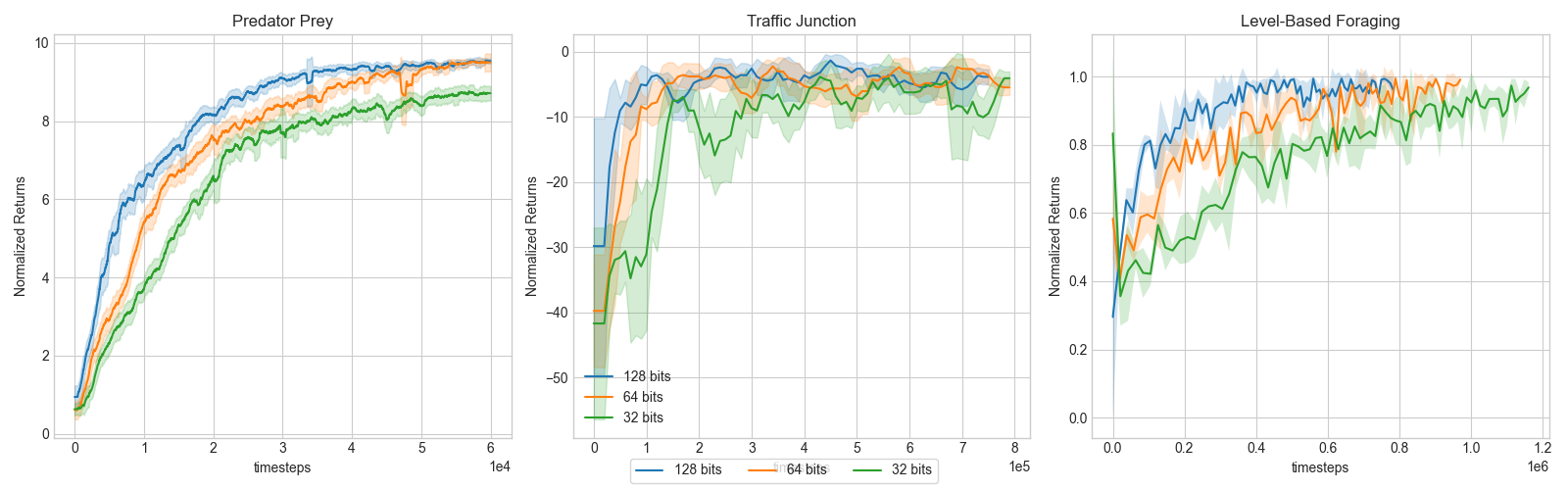}
    \vspace{-0.7cm}
    \caption{Learning curves with different bits (128,64,32) respectively in three environments. This figure demonstrates that the bit-length (capacity) of the communication vector -- which represents the amount of information it can carry -- has distinct influence on learning speed. The size of this influence varies across different tasks. }
    \label{fig:diff_bits}
\end{figure*}

%Diff Env
The multi-agent traffic junction environment also has three versions with incremental levels. Table \ref{tab:Diff_Traffic_Junction_Settings} lists the details. 
Figure \ref{fig:traffic_junction_diff_env} shows that the parameters in Table \ref{tab:Diff_Traffic_Junction_Settings} pulls down the converged total reward and brings more variance into cooperation among agents.
\begin{table}[]
    \centering
    \begin{adjustbox}{width=\columnwidth,center}
    \begin{tabular}{c|c|c|c}
                & Agents Num & Grid Size & Max Steps \\
    Easy Version & 5 & 6 & 20  \\
    Medium Version & 14 & 10 & 40  \\
    Hard Version & 20 & 18 & 80\\
    \end{tabular}
    \end{adjustbox}
    \caption{Settings of different versions TrafficJunction environment}
    \label{tab:Diff_Traffic_Junction_Settings}
\end{table}
% \begin{figure}[h]
%     \centering
%     \includegraphics[scale=0.55]{figures/traffic_junction_baseline_compare.png}
%     \vspace{-0.7cm}
%     \caption{The curves show different difficulties has influence on the training.}
%     \label{fig:traffic_junction_baselines}
% \end{figure}
%Diff Comm
Same as in the Predator-Prey environment, we want to test our algorithm on different bits of message on the Traffic Junction environment. We can tell from the centre plot of  Figure \ref{fig:diff_bits} that, unlike the predator-prey environment, the bit-length of information causes disparity in the final performance; in a traffic junction environment, it primarily affects the learning speed.

\begin{figure}[h]
    \centering
    \includegraphics[scale=0.55]{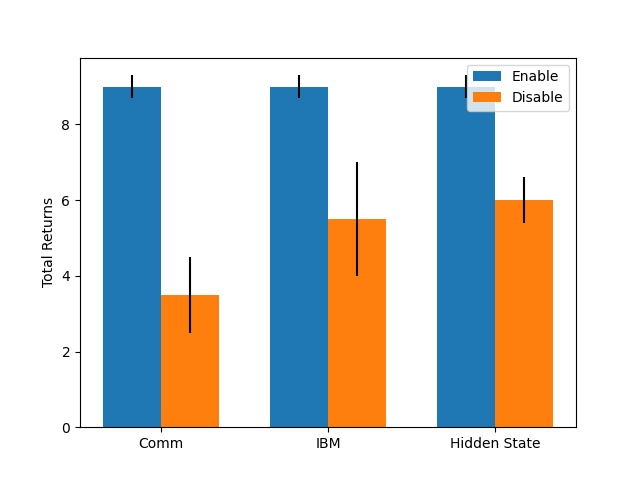}
    \vspace{-0.7cm}
    \caption{The blue bars are all function enabled IEC with communication vector of size 128 bits. The orange bar is the returns the algorithm can achieve after disabling  the corresponding feature.}
    \label{fig:ablation_study}
\end{figure}

\subsection{Level-based Foraging}
Level-based Foraging (LBF) is a multi-agent environment focusing on the coordination of the agents involved \cite{papoudakis2021benchmarking}. Compared to the PredatorPrey environment, agents and food in this world are assigned within a level; only when the agent's level is equal to or higher than a certain limit,  can it successfully collect the food, as in Figure \ref{fig:envs}. The action space contains four directional moves and a loading action. This environment is more challenging than previous ones due to its additional limitations.

Figure \ref{Level-based Foraging} shows our algorithm obtains higher learning speed and normalised returns than QMIX and VDN. The constraint that only a certain level can elicit pick up of the food decreases the probability that the agents randomly hit the target. This also requires richer observation and more efficient cooperation, like not wasting time collecting the same food. In such a scenario, we can see our intention encoded communication plays a vital role in assisting agents in achieving their goals.

\subsection{Ablation Study}
This section discusses some key building modules communication, IBM, and hidden state, respectively. They have a significant influence on final performance. We conducted an ablation study by solely disabling these features compared with the fully functional IEC in the PredatorPrey environment. The difficulty level is easy, and the message length stays at 128 bits.

\textbf{No Communication}
We analyse the functionality of the communication mechanism. In this setting, agents do not have any exchanges, which means the IBM is also unavailable due to the lack of passing messages; they can only use their observation as input for decision making. 
% \begin{figure}[h]
%     \centering
%     \includegraphics[scale=0.55]{figures/comm_on_off.png}
%     \caption{The total reward while training along the episodes. The environment is a PredatorPrey Env. There are 3 predators and 1 prey, the prey is fixed and predators are randomly spawned in the grids. The yellow curve is with communication enabled and the blue is disables. It's clear that without communication, agents can't reach the same high scores.}
%     \label{fig:comm_functionality}
% \end{figure}
% We run the experiments with the gate open or close representing whether the agents communicate or not.
Figure \ref{fig:ablation_study}'s first pair shows that the predator
% are not able to learn cooperation behaviour and 
encounters greater total reward loss than the method with fully functional communication.

\textbf{No IBM}
For validation of our Inferring Belief Module (IBM)'s functionality, we run one experiment that all predators only receive others' hidden state as a message, i.e., agents do not exchange each other's embedded intentions during the experiment. Figure \ref{fig:ablation_study}'s centre plot shows that without the extra belief inference of other agents' messages cause about $38\%$ performance decrease.

% \begin{figure}[h]
%     \centering
%     \includegraphics[scale=0.55]{figures/comm_with_with_without_vae_predictions.png}
%     \caption{The blue curve is the messages incorporating the agent modeling information. The knowledge of other agents benefits the cooperation and lead to higher total reward at the end.}
%     \label{fig:modeling_functionality}
% \end{figure}

\textbf{No Hidden State}
A hidden state is the vector we use to embed the agent's previous state using an RNN. Details can be found in Section \ref{subsec:message_generation_module}. As the right plot of Figure \ref{fig:ablation_study} shows, the algorithm reaches around 2/3 returns compared to IEC if the hidden state is missing in a transmitted message. We can see that the compact information of the agent's state also brings some help to complete the cooperation tasks. But without IBM, it can't reach maximal returns.

\section{Conclusion}
This paper proposes IEC framework that can mimic the multiple agent's natural communication in cooperative tasks. We devise two modules: Infer Beliefs Module (IBM) and Message Generation Module. These modules abstractly represent functional areas in the brain for successful comprehension inference and grammar generation. We also adopt the idea that agents learn the communication and underlying grammar by trial and error (co-evolution). We empirically demonstrated that our algorithm could outperform other multi-agent baselines with a significant gain in data efficiency. Such an algorithm stays robust under various environment settings, like different difficulties and message length. The ablation study also investigates the processes' key factors and reveals their contribution. In future work, we would like to explore more generative structures like GANs for inferring belief modules. 
% It would also be interesting to deploy our method in real-world robot fleet.

% In the unusual situation where you want a paper to appear in the
% references without citing it in the main text, use \nocite
\nocite{langley00}

\bibliography{example_paper}
\bibliographystyle{icml2021}

\end{document}